\pdfoutput=1

\documentclass[11pt]{article}

\usepackage[final]{ACL2023}

\usepackage{times}
\usepackage{latexsym}

\usepackage{booktabs, multirow} 
\usepackage{soul}
\usepackage{xcolor,colortbl} 
\usepackage{changepage,threeparttable} 
\usepackage{graphicx}
\usepackage{makecell}
\usepackage{amsmath}
\usepackage{amsfonts}
\usepackage{mathtools}
\usepackage{subcaption}
\usepackage[T1]{fontenc}

\usepackage[utf8]{inputenc}
\usepackage{multirow,graphicx,rotating}
\newcommand{\std}[1]{{\scriptsize$\pm$#1}} 
\usepackage{microtype}
\usepackage{stfloats}   

\usepackage{inconsolata}

\title{Retrieval Disagreement as a Signal: Diagnosing and Steering\\ Multilingual Dense Embedders}

\author{
Faeze Ghorbanpour\qquad Shaghayegh Kolli\qquad Jana Diesner\qquad Alexander Fraser \vspace{.2cm}\\
School of Computation, Information and Technology, TU Munich \\
Munich Center for Machine Learning (MCML)\\
\vspace{.1cm}
{\tt \small \{faeze.ghorbanpour, shaghayegh.kolli\}@tum.de}\\
}

\begin{document}
\maketitle
\begin{abstract}

Multilingual dense retrievers place all languages in one shared space so that a question finds relevant documents, regardless of which language they are written in. 
We show that the same question, asked in different languages, retrieves substantially different documents, and this disagreement is not because of weak language understanding. 
Each language retrieves well on its own and ranks candidates consistently with the other languages; divergence appears only when languages share an index. 
We trace the mechanism: each language keeps its own ranking, but in the shared index, its correct answers are pushed far down, buried under non-relevant documents that are either in the query's own language or in English, depending on the model.
Because this disagreement is structured, it can also serve as a correction signal. 
We introduce disagreement-guided steering: for each language, we contrast the documents that other languages agree on for a question against those that language retrieves idiosyncratically, and take the difference of their representations as a language-specific direction, added to that language's queries at inference. 
Against baselines of language-mean removal, subspace removal, and per-language rotation, it most strongly reduces same-language crowding while remaining competitive on recall, trading off against mean removal, which better preserves ranking quality.

\end{abstract}

\section{Introduction}
Multilingual dense retrievers map queries and documents from many languages into one vector space, so that relevance rather than language decides what is retrieved \citep{wang2024multilinguale5textembeddings, qwen3embedding}. The intent is met unevenly. The same model retrieves well for some query languages and fails badly for others, as documents in the query's own language, or in English, fill the top ranks and push relevant foreign-language documents out \citep{yang2025languagebiasinformationretrieval, wang2026languagesmatterunderstandingmitigating,  amiraz-etal-2025-cross}. What separates the languages that fail from those that succeed is not visible in their scores.

Scoring a language against gold labels shows how often it retrieves the right documents, but not why it misses them. To see why, we ask the same question in several languages and compare what each retrieves.
The retrieved sets diverge, and how they diverge is diagnostic. 
We call this cross-lingual disagreement and make it our object of study. 
It requires no relevance labels and characterizes failed language by its departure from successful languages rather than by a single number.


Disagreement is diagnostic because it separates two explanations that a single score cannot. A language may fail because the model represents it poorly, or because the languages interfere only once they share an index. The two call for different fixes, better training in the first case, a geometric correction in the second, but from gold scores alone, they look identical. Prior work does not separate them: monolingual and cross-lingual retrieval are typically measured on different collections, confounding the effect of mixing with a change of items, while audits quantify how unequal the mixed ranking is \citep{wang2026languagesmatterunderstandingmitigating}, and mitigations act on the output space, transforming the index \citep{jang2026shift}, removing language identity post hoc \citep{kim2026langsae}, projecting out a language subspace \citep{xie-etal-2022-discovering}, or correcting hubness \citep{lample2018word, anonymous2026islands}. Each presupposes which explanation holds rather than testing it.

Disagreement can be read at three points, each answering a different question. 
Among languages retrieving only within their own documents, it asks whether they behave alike in isolation. They do: every language retrieves competently and orders concepts much as the others do, so representational quality is not the binding constraint. 
Between monolingual and mixed retrieval, it asks what sharing an index changes. The answer is composition rather than order: each language keeps its internal ranking, while its documents drop far down the combined list, displaced by same-language non-relevant material rather than by globally popular hub documents \citep{lample2018word, anonymous2026islands}. 
Among languages competing inside the mixed index, it asks whether they converge. They do not, and the divergence follows two regimes: some models crowd a query with its own language, while others pull the ranking toward English. 
The pattern is directional: a language's documents stay reachable by other languages' queries, while its own queries retrieve almost exclusively within that same language~\citep{lim2025language}.

Disagreement is also actionable. Knowing which documents other languages retrieve and which they miss gives not only the size of the gap but also a direction across it. 
We introduce disagreement-guided steering: contrast the documents other languages agree on for a question against those a language retrieves idiosyncratically, and add the difference of their representations to that language's queries at inference. 
This requires no neuron attribution, no training, and no relevance labels; its directions are estimated on a separate corpus, never on the evaluation data. We compare it against per-language mean removal \citep{libovicky-etal-2020-language}, language-subspace removal \citep{xie-etal-2022-discovering}, and rotation toward a cross-lingual consensus \citep{liu2019investigating}. The corrections are separated by what they repair: mean removal gives the best ranking quality and rank agreement, while steering gives the largest reduction in same-language crowding and remains competitive on recall, and holds up on non-parallel corpora, where documents are not translations of one another.

Our contributions are: \textbf{(1)} Cross-lingual disagreement as a label-free diagnostic, read within languages, across the monolingual-to-mixed transition, and among languages sharing an index. \textbf{(2)} An anatomy of cross-lingual retrieval failure: demotion rather than loss of correct answers, displacement by same-language non-relevant documents rather than hubs, two model-dependent crowding regimes, and a directional asymmetry between documents and queries. \textbf{(3)} Disagreement-guided steering, an inference-time correction derived from retrieval disagreement, requiring no training and no access to the evaluation corpus.~\footnote{Code, scripts, and the estimation pools will be released publicly upon acceptance.}

\section{Related Work}

\paragraph{Geometry of Multilingual Embedding Spaces}
A long line of work characterizes multilingual embedding spaces through their geometry and corrects them accordingly. \citet{lample2018word} showed that a few hub vectors become nearest neighbors of almost everything in high-dimensional spaces and proposed CSLS as a correction, which later work extended with normalization steps that strip language-specific structure before alignment \citep{ruder2019survey, aboagye2022normalization, pan2025hubness}. A second strand targets language identity directly: \citet{roy2020lareqa} and \citet{reimers-gurevych-2020-making} showed that embeddings clustering by language causes retrievers to prefer irrelevant same-language documents over relevant foreign ones, \citet{libovicky-etal-2020-language} and \citet{yang2021simple} removed this bias by subtracting per-language means, and \citet{xie-etal-2022-discovering} showed that language identity occupies a low-rank subspace
that can be projected away. Orthogonal Procrustes alignment provides a third correction, long used to relate separately trained embedding spaces \citep{liu2019investigating, lample2018word}. We evaluate all three under identical conditions and against a correction that, rather than assuming a fixed geometric form, derives its direction from retrieval behavior.



\paragraph{Language Bias in Multilingual Retrieval}
Language bias in retrieval has been documented and mitigated at the level of the final ranking. \citet{yang2025languagebiasinformationretrieval} showed that multilingual retrievers treat query languages unequally and proposed a training objective to reduce the gap. \citet{wang2026languagesmatterunderstandingmitigating} found that multilingual RAG pipelines systematically promote English and query-language documents. Mitigations transform the document index \citep{jang2026shift} or edit language identity out of the output embeddings \citep{kim2026langsae}. 
All of these measure a language against gold labels: they establish that some languages retrieve worse than others, but not why, since a per-language score never compares one language's retrievals against another's.
We measure languages against one another instead, which requires no relevance judgments and distinguishes retrieving different material from retrieving the same material in a worse position. 

\paragraph{Steering and Model Interventions}
Modifying model weights at inference has become a standard alternative to fine-tuning, both for controlling generation \citep{turner2024steeringlanguagemodelsactivation, rimsky-etal-2024-steering} and for improving multilingual behavior. \citet{sundar-etal-2025-steering} identifies expert neurons and modulates them to increase cross-lingual alignment in multilingual language models, reporting gains on cross-lingual retrieval, while \citet{10.1145/3626772.3657819} steers activations in a generative reader within a CLIR pipeline, and \citet{pokharel2026crosslingualactivationsteeringmultilingual} selectively modulates neuron activations for multilingual classification and generation. These approaches locate the intervention through attribution, by identifying which units encode a property, and require an anchor language or a target behavior to steer toward. Our construction differs in the source of the direction: we obtain it from the retrieval behavior of the languages themselves. 

\section{Methodology}
\label{sec:approach}
Multilingual retrievers are trained so that a query in any language reaches the relevant documents in every other language. Whether they do is normally judged one language at a time, by scoring each ranking against relevance labels. Such a score establishes that some languages are served worse than others, but it does not reveal the character of the loss: a language may retrieve unrelated material, or the right material only in its own language, or the right material demoted beneath documents no other language would have chosen. These are distinct failures with distinct remedies, and a single number cannot separate them.

We therefore evaluate the rankings against each other rather than against labels alone. On parallel collections, one information need exists in every language, and each passage exists as translations that we treat as one item, so the rankings produced by different languages become directly comparable. Their pairwise relation, which we call cross-lingual agreement, is measured at three points that isolate successive stages of retrieval: within languages restricted to their own documents, where competition is absent; between a language's restricted and shared-index rankings, where the only change is the presence of other languages; and among languages inside the shared index, where they compete directly.


Disagreement also shows how to fix a failing language. For a question, the other languages find documents that this language misses, and it retrieves documents they avoid. Subtracting the average of the second group from the average of the first, in embedding space, gives a direction that points from where this language's queries land toward where the others' do. We average this over questions to get one direction per language, weighting questions by how much the language disagrees on them, and add it to its queries at retrieval time. We steer queries only, leaving document embeddings untouched. This keeps the method inference-only and requires no re-encoding or re-indexing of the collection: the correction is applied per query at search time, so it can be dropped into an existing index without rebuilding it.

\subsection{Setup and Notation}
\label{sec:setup}

Let $\mathcal{L}$ be the language set and $\mathcal{D}$ the index, with
$\mathcal{D}_\ell$ its language-$\ell$ subset. Document $d$ has semantic identity $g(d)$, shared with its translations, and need $x$ appears as query $q^x_\ell$ in every language. We write $e(\cdot)\in\mathbb{R}^n$ for the final-layer embedding and score by cosine
similarity. Retrieval returns the top $K=20$ documents as identities,
\begin{equation}
\resizebox{.9\columnwidth}{!}{$
R_\ell(x)=\operatorname*{top\text{-}\mathit{K}}_{d\in\mathcal{D}_\ell} g(d),
\quad
\tilde{R}_\ell(x)=\operatorname*{top\text{-}\mathit{K}}_{d\in\mathcal{D}} g(d),
$}
\end{equation}
so $R_\ell$ is monolingual and $\tilde{R}_\ell$ shared-index retrieval. Interventions act
on $e(\cdot)$ at the final layer, estimated on an external corpus $\mathcal{C}$ disjoint
from $\mathcal{D}$, with $\mu_\ell$ the mean of $\mathcal{C}_\ell$.

\subsection{Agreement}
\label{sec:agreement}

For ranked identity lists $P,Q$ we measure set and rank agreement,
\begin{align}
A_{\mathrm{set}}(P,Q) &= \frac{|P\cap Q|}{|P\cup Q|},\\
A_{\mathrm{rank}}(P,Q) &= \rho\!\left(\pi_P(P\cap Q),\pi_Q(P\cap Q)\right),
\end{align}
with $\pi_P(\cdot)$ an item's position in $P$ and $\rho$ Spearman correlation. The three
readings apply these to $\big(R_\ell(x),R_m(x)\big)$,
$\big(R_\ell(x),\tilde{R}_\ell(x)\big)$ and
$\big(\tilde{R}_\ell(x),\tilde{R}_m(x)\big)$, averaged over needs $x$ and languages
$m\neq\ell$. We also record the share of $\tilde{R}_\ell$ contributed by
$\mathcal{D}_\ell$ and $\mathcal{D}_{\mathrm{en}}$, and the rank of the correct
identity's foreign-language copies.

\subsection{Interventions}
\label{sec:interventions}

\label{sec:existing}

\noindent\textbf{Mean removal} subtracts the language offset and renormalizes,
$e'(x)=\big(e(x)-\mu_{\ell(x)}\big)/\lVert e(x)-\mu_{\ell(x)}\rVert$.

\noindent\textbf{Subspace removal} stacks the centred
means as $M=[\mu_1-\bar{\mu},\dots,\mu_{|\mathcal{L}|}-\bar{\mu}]$ and projects out its
top $r$ left singular vectors $U_r$,
\begin{equation}
e'(x)=\left(I-U_rU_r^{\top}\right)e(x).
\end{equation}

\noindent\textbf{Rotation} aligns each language to the consensus
$\bar{e}(g)=\frac{1}{|\mathcal{L}|}\sum_m e(d^g_m)$ by Procrustes, applying the map to
that language's queries,
\begin{equation}
W_\ell=\operatorname*{arg\,min}_{W^{\top}W=I}\sum_g
\big\lVert We(d^g_\ell)-\bar{e}(g)\big\rVert^2 .
\end{equation}

\noindent \textbf{Disagreement-guided steering}
\label{sec:ours}
Let $C_\ell(x)$ be the consensus of the other languages, obtained by reciprocal-rank fusion over $\{R_m(x)\}_{m\neq\ell}$ on $\mathcal{C}$. We contrast the identities the consensus retrieves but $\ell$ misses, $P_\ell(x)=C_\ell(x)\setminus R_\ell(x)$, against those $\ell$ retrieves alone, $N_\ell(x)=R_\ell(x)\setminus C_\ell(x)$,
\begin{equation}
u_\ell(x)=\frac{1}{|P_\ell(x)|}\!\!\sum_{g\in P_\ell(x)}\!\! e(d^g)
-\frac{1}{|N_\ell(x)|}\!\!\sum_{g\in N_\ell(x)}\!\! e(d^g_\ell),
\end{equation}
where $d^g$ is concept $g$'s document in a consensus language that retrieves it (the first such language, when several do), and $d^g_\ell$ is that concept's document in language $\ell$ itself. We then aggregate with weights $w_\ell(x)=1-A\big(R_\ell(x), C_\ell(x)\big)$,
\begin{equation}
v_\ell=\frac{\sum_x w_\ell(x)\,u_\ell(x)}{\sum_x w_\ell(x)} .
\end{equation}
where $A(\cdot,\cdot)$ is the set (Jaccard) overlap between the two top-$K$ lists, so a query counts more when $\ell$ and the consensus disagree more.

At inference, the direction is added to the language's queries, scaled by the query norm so $\alpha$ transfers across models,
\begin{equation}
e'(q_\ell)=e(q_\ell)+\alpha\lVert e(q_\ell)\rVert\,v_\ell/\lVert v_\ell\rVert .
\end{equation}
Positives come from other languages' documents and negatives from $\ell$'s own, so the direction points away from what a language over-retrieves and toward what the others agree on. Nothing is trained, no relevance judgments are used, and $\mathcal{C}$ is disjoint from the evaluation index.~\footnote{Equations~(6)--(8) are a motivated construction, not an optimality guarantee: the direction is validated empirically rather than derived as provably optimal.}

\section{Experimental Setup}

\paragraph{Datasets.}
We evaluate three multilingual retrieval collections with complementary structures. Belebele \citep{bandarkar-etal-2024-belebele} and XQuAD
\citep{Artetxe:etal:2019} are fully parallel: every passage exists in all languages and shares a semantic identity, so retrieved sets are directly
comparable across languages and disagreement can be measured directly. Belebele spans ten languages across five scripts, XQuAD six. Multi-EuP
\citep{yang-etal-2023-multi-eup} is included for contrast: drawn from real European Parliament speeches, its language pools are strongly imbalanced, and its relevance comes from human-assigned subject labels rather than translation identity, so documents do not correspond one-to-one across languages. It tests whether findings from clean parallel conditions survive in a realistic index. All documents are passage-level, so single-vector retrieval is not confounded with long-document aggregation. Appendix~\ref{app:datasets} gives languages, access, and licensing.

\paragraph{Models.}
We evaluate seven multilingual embedding models. mE5-base and mE5-large \citep{wang2024multilinguale5textembeddings} are encoder-based retrievers used with their official \textit{query:} and \textit{passage:} prefixes and mean pooling. BGE-M3 \citep{bge-m3} is a further encoder-based retriever, evaluated in its single-vector dense mode with CLS pooling and no prompt, as its training prescribes. Qwen3-Embedding-0.6B and Qwen3-Embedding-8B \citep{qwen3embedding} are decoder-based, read out at the final token, and included as a matched pair so that scale can be varied within one family. e5-mistral (e5-mistral-7b-instruct) is a decoder-based instruction embedding model~\citep{wang2023improving, wang2022text}. It uses last-token pooling and is less cleanly multilingual because its backbone is English-centric. EmbeddingGemma-300M \citep{vera2025embeddinggemmapowerfullightweighttext} is a compact decoder-based model with mean pooling and task-prefixed inputs, whose two projection heads we apply so that its final representation matches the deployed embedding. Every model is used in inference mode without training, and each is evaluated only in the pooling and prompt configuration its documentation specifies. 
The details regarding the model and experimental procedures can be found in Appendix~\ref{app:models}.



\paragraph{Estimation corpora.}
All interventions require language-level statistics.  
Estimating these on the evaluation index would make the corrections transductive, so we estimate them on a separate corpus. 
We use FLORES-200 \citep{nllbteam2022languageleftbehindscaling} for XQuAD and Multi-EuP, as they have no shared content with it. Belebele, however, is built from FLORES passages; so we estimate Belebele's directions on TED2020 \citep{ted_talk}, which shares no content with Belebele. The steering scale $\alpha$ is selected on the estimation corpus and held fixed across datasets; only the per-model value varies. 

\paragraph{Evaluation Metrics.}
We measure retrieval quality with Recall@20 and nDCG@20 at the final layer, in two variants that differ only in which gold copies count. All-gold nDCG/Recall count every copy of the answer, including the same-language one; we use these in the diagnosis to characterize the general effect of mixing. Cross-lingual nDCG/Recall count only foreign-language copies, excluding the same-language answer from both the hits and the ideal ranking; we use these for the interventions, to isolate the cross-lingual failure the corrections target. We also report the share of a query’s top-20 in its own language and in English, which separates self-crowding from English-gravity, and cross-lingual disagreement, the overlap and rank correlation between the rankings two languages produce for the same question, averaged over all other languages. Together, these capture whether a correction retrieves more relevant documents, changes which languages fill the ranking, and brings languages into agreement.

\section{Experimental Results}




\subsection{Diagnosis}
\label{sec:diagnosis}

We read cross-lingual disagreement at the three points defined in Section~\ref{sec:agreement}, moving from each language in isolation to the languages competing in a shared index. Table~\ref{tab:isolation} reports the first reading, Table~\ref{tab:mixing} the second and third.

\begin{table}[t]
\centering
\small
\setlength{\tabcolsep}{4pt}
\begin{tabular}{p{1.7cm}p{0.8cm}p{0.7cm}p{0.5cm}p{0.8cm}p{0.7cm}p{0.5cm}}
\toprule
& \multicolumn{3}{c}{Belebele} & \multicolumn{3}{c}{XQuAD} \\
\cmidrule(lr){2-4}\cmidrule(lr){5-7}
Model & nDCG$_\text{all}$ & Rec$_\text{all}$ & Agr & nDCG$_\text{all}$ & Rec$_\text{all}$ & Agr \\
\midrule
mE5-large   & 95.0 & 99.0 & 53.6 & 97.2 & 99.9 & 66.3 \\
mE5-base    & 93.5 & 98.6 & 50.7 & 96.1 & 99.8 & 63.1 \\
bge-m3      & 91.1 & 97.6 & 53.8 & 90.3 & 98.0 & 62.5 \\
Qwen3-0.6B  & 87.7 & 95.5 & 50.8 & 88.4 & 97.6 & 59.2 \\
Qwen3-8B    & 92.7 & 98.1 & 52.5 & 90.8 & 99.0 & 59.9 \\
e5-mistral  & 91.8 & 97.7 & 52.1 & 94.6 & 99.2 & 62.6 \\
Gemma  & 69.8 & 87.2 & 35.0 & 76.0 & 92.3 & 45.7 \\
\bottomrule
\end{tabular}
\caption{Isolation: monolingual nDCG@20, Recall@20, and among-language agreement (Agr) when each language retrieves only within its own documents. 
}
\label{tab:isolation}
\end{table}

\noindent
\textbf{Languages are strong in isolation.}
When each language is retrieved only from its own documents, every model performs well, with only minor variation across languages. On both
collections, monolingual nDCG@20 ranges from $88$ to $97$ and recall from $96$ to $100$ across every language, for all models but the compact
Gemma-300M; cross-language rank agreement is moderate and roughly uniform ($50$--$66$). Gemma-300M is the lone exception (nDCG $70$/$76$, recall $87$/$92$, agreement $0.35$/$0.46$): as the smallest model and a general-purpose rather than retrieval-specialized embedder, it produces
weaker and less consistent rankings even before competition. The difference that later separates a failing language from a succeeding one is therefore not a difference in how well the model represents it in isolation, where every language is retrieved well, but something that emerges under competition.

\begin{table*}[t]
\centering
\small
\setlength{\tabcolsep}{4pt}
\begin{tabular}{l ccccc ccccc}
\toprule
& \multicolumn{5}{c}{Belebele} & \multicolumn{5}{c}{XQuAD} \\
\cmidrule(lr){2-6}\cmidrule(lr){7-11}
Model & nDCG$_\text{all}$ & Rec$_\text{all}$ & Gold & Ord & Agr & nDCG$_\text{all}$ & Rec$_\text{all}$ & Gold & Ord & Agr \\
\midrule
bge-m3      & 87.5 & 96.0 &  55 & 61.0 & 51.9 & 84.9 & 96.5 &  25 & 67.2 & 60.9 \\
mE5-base    & 69.0 & 84.3 & 144 & 62.1 & 30.4 & 72.6 & 91.2 &  42 & 67.8 & 39.9 \\
mE5-large   & 73.7 & 88.9 &  96 & 70.2 & 36.4 & 76.7 & 94.0 &  31 & 75.6 & 45.4 \\
Qwen3-0.6B  & 80.7 & 91.3 & 108 & 55.7 & 41.5 & 79.6 & 94.5 &  34 & 61.5 & 53.7 \\
Qwen3-8B    & 88.7 & 96.0 &  50 & 60.4 & 36.8 & 85.0 & 97.1 &  20 & 67.0 & 49.8 \\
e5-mistral  & 60.1 & 72.3 & 240 & 66.4 & 35.5  & 60.5 & 74.1 & 113 & 70.9 & 50.3  \\
Gemma  & 60.9 & 78.9 & 262 & 37.2 & 25.9 & 67.9 & 87.1 &  74 & 49.4 & 41.7 \\
\bottomrule
\end{tabular}
\caption{Mixing: mixed-index nDCG@20, Recall@20, foreign-gold rank
(Gold), preservation of each language's monolingual order (Ord), and among-language
agreement inside the shared index (Agr). 
}
\label{tab:mixing}
\end{table*}

\noindent
\textbf{Mixing demotes rather than disorders.}
Placing all languages in one index changes retrieval sharply, but not by scrambling them. Comparing each language's monolingual ranking with its ranking in the shared index (Table~\ref{tab:mixing}), the internal order is largely preserved, between $56$ and $76$ on the parallel collections, while the correct foreign-language documents fall far down the list: their rank drops from the top to between $50$ and $240$ depending on the model; these documents reappear at larger retrieval cutoffs, confirming they are demoted rather than lost (Appendix~\ref{app:k100}).
The relevant documents are still ordered coherently relative to one another; they are simply buried beneath other material. 

\noindent
\textbf{Cross-lingual retrieval works before competition.} Table~\ref{tab:isolation} shows each language retrieves well within its own documents, but not that it retrieves \emph{foreign}-language documents well in isolation. We therefore test each query language against one foreign document-language pool at a time, before mixing all languages together. Retrieval is near-perfect in this bilingual setting: Recall@20 averages $97.8$ (mE5-large), $92.8$
(Qwen3-0.6B), and $97.6$ (Qwen3-8B) across every language pair. The same queries and documents fail only once all languages compete in the shared index (Table~\ref{tab:mixing}). A representational deficit would show up in the bilingual pools too; it does not, so the failure is interference under competition, not an inability to represent foreign documents.


\begin{table*}[t]
\centering
\small
\setlength{\tabcolsep}{4pt}
\begin{tabular}{l ccccc ccccc}
\toprule
& \multicolumn{5}{c}{Belebele} & \multicolumn{5}{c}{XQuAD} \\
\cmidrule(lr){2-6}\cmidrule(lr){7-11}
Model & Same & En & Other & Hub & Drop & Same & En & Other & Hub & Drop \\
\midrule
mE5-large   & 87.7 &  0.1 &  3.8 & 8.5 & 23/24 & 89.6 &  0.2 &  5.1 & 8.7 & 22/24 \\
mE5-base    & 77.7 &  0.5 &  9.7 & 7.3 & 11/12 & 79.5 &  1.1 & 11.2 & 7.1 & 11/12 \\
e5-mistral  & 91.6 &  0.0 &  5.6 & 3.4 & 29/32 & 97.3 &  0.0 &  1.6 & 2.6 & 29/32 \\
bge-m3      &  8.7 &  6.6 & 72.8 & 8.5 & 22/24 & 13.9 & 11.7 & 67.7 & 8.1 & 23/24 \\
Qwen3-8B    & 38.4 &  9.7 & 33.3 & 7.5 & 34/36 & 43.5 & 12.4 & 31.3 & 5.8 & 30/36 \\
Qwen3-0.6B  & 22.8 & 19.0 & 47.2 & 7.9 & 25/28 & 26.2 & 16.6 & 50.5 & 2.6 & 23/28 \\
Gemma  & 22.8 & 13.8 & 47.9 & 8.9 & 19/24 & 12.8 & 27.3 & 50.1 & 6.2 & 19/24 \\
\bottomrule
\end{tabular}
\caption{Composition of the non-relevant documents that replace a demoted cross-lingual gold, as a percentage of all entrants: same-language (Same), English (En), and other foreign languages (Other), with the hub share (Hub) and the average drop layer (Drop). 
}
\label{tab:replacement}
\end{table*}

\begin{table*}[tb!]
\centering
\footnotesize
\setlength{\tabcolsep}{2.0pt}
\begin{tabular}{cl p{1cm}p{1cm}p{1cm}l p{1cm}p{1cm}p{1cm}l p{1cm}p{1cm}p{1cm}l}
\toprule
& & \multicolumn{4}{c}{Belebele} & \multicolumn{4}{c}{XQuAD} & \multicolumn{4}{c}{Multi-EuP} \\
\cmidrule(lr){3-6}\cmidrule(lr){7-10}\cmidrule(lr){11-14}
& Method & Rec$_\text{xl}$ & nDCG$_\text{xl}$ & Own & En & Rec$_\text{xl}$ & nDCG$_\text{xl}$ & Own & En & Rec$_\text{xl}$ & nDCG$_\text{xl}$ & Own & En \\
\midrule
\multirow{7}{*}{\rotatebox[origin=c]{90}{mE5-large}}
 & baseline  & 85.0 & 45.7 & 69.3 & 8.5 & 87.7 & 47.0 & 71.3 & 14.0 & 76.8 & 23.0 & 45.2 & 35.3 \\
 & random    & 85.0\std{0.2} & 45.7\std{0.2} & 69.3\std{0.2} & 8.5\std{0.1} & 87.7\std{0.2} & 47.1\std{0.1} & 71.3\std{0.2} & 14.1\std{0.1} & 76.7\std{0.2} & 22.9\std{0.1} & 45.4\std{0.3} & 35.0\std{0.3} \\
 & permuted  & 33.2\std{0.7} & 6.5\std{0.2} & 96.1\std{0.1} & 10.0 & 17.4\std{0.3} & 2.8\std{0.1} & 98.6\std{0.0} & 16.4\std{0.0} & 9.5\std{0.1} & 1.3\std{0.0} & 98.2\std{0.0} & 5.4\std{0.0} \\
 & subspace  & 93.5\std{0.7} & 63.4\std{0.8} & 44.9\std{1.9} & 8.0\std{0.4} & 96.1\std{0.3} & 59.4\std{0.6} & 55.4\std{1.0} & \textbf{11.8\std{0.2}} & 79.8\std{1.1} & 27.2\std{1.3} & 37.6\std{4.0} & 29.1\std{2.8} \\
 & rotate    & 83.5\std{2.2} & 45.3\std{3.7} & \textbf{22.2\std{7.6}} & 19.4\std{8.9} & 97.9\std{0.3} & 68.5\std{1.9} & 21.9\std{3.8} & 34.6\std{6.0} & 68.0\std{2.3} & 15.4\std{1.4} & \textbf{8.4\std{1.3}} & 70.4\std{9.8} \\
 & centering & 93.3\std{0.2} & 56.7\std{0.4} & 44.1\std{0.7} & \textbf{7.1\std{0.2}} & 98.7\std{0.0} & 67.6\std{0.1} & 36.9\std{0.2} & 15.6\std{0.1} & 78.6\std{0.2} & 31.6\std{0.1} & 20.4\std{0.2} & \textbf{21.3\std{0.1}} \\
 & steering  & \textbf{96.9\std{0.3}} & \textbf{64.5\std{0.2}} & 22.9\std{1.7} & 10.2\std{0.2} & \textbf{99.3\std{0.0}} & \textbf{72.0\std{0.1}} & \textbf{14.5\std{0.1}} & 15.7\std{0.0} & \textbf{85.0\std{0.1}} & \textbf{32.1\std{0.0}} & 8.6\std{0.1} & 53.5\std{0.2} \\
\midrule
\multirow{7}{*}{\rotatebox[origin=c]{90}{mE5-base}}
 & baseline  & 84.2 & 40.4 & 70.4 & 8.5 & 84.5 & 41.5 & 72.3 & 14.2 & 65.1 & 16.3 & 55.4 & 30.1 \\
 & random    & 84.2\std{0.3} & 40.3\std{0.2} & 70.4\std{0.3} & 8.5\std{0.1} & 84.4\std{0.2} & 41.4\std{0.2} & 72.4\std{0.2} & 14.2\std{0.1} & 65.0\std{0.3} & 16.2\std{0.2} & 55.5\std{0.6} & 29.7\std{0.5} \\
 & permuted  & 13.8\std{0.5} & 2.2\std{0.1} & 98.7 & 10.1 & 9.1\std{0.2} & 1.2\std{0.0} & 99.4\std{0.0} & 17.0\std{0.0} & 11.7\std{0.2} & 1.6\std{0.0} & 89.1\std{0.3} & 12.2\std{0.2} \\
 & subspace  & 91.8\std{1.0} & 55.8\std{1.7} & 45.5\std{3.7} & \textbf{5.7\std{0.9}} & 93.9\std{0.6} & 53.2\std{1.5} & 50.5\std{2.2} & \textbf{8.8\std{1.1}} & 67.2\std{4.0} & 18.0\std{2.2} & 47.4\std{7.9} & 17.4\std{7.5} \\
 & rotate    & 70.9\std{3.0} & 29.9\std{2.6} & 36.0\std{7.6} & 18.7\std{6.3} & 90.5\std{1.6} & 50.3\std{2.2} & 45.2\std{5.3} & 14.1\std{2.0} & 66.1\std{1.7} & 16.6\std{0.8} & \textbf{15.4\std{2.3}} & 48.8\std{9.4} \\
 & centering & 85.1\std{0.6} & 40.0\std{0.6} & 60.1\std{1.0} & 9.3\std{0.2} & 93.0\std{0.2} & 51.9\std{0.2} & 48.0\std{0.3} & 21.8\std{0.2} & 53.6\std{0.2} & 12.0\std{0.1} & 24.7\std{0.3} & \textbf{12.0\std{0.2}} \\
 & steering  & \textbf{93.3\std{0.1}} & \textbf{56.1\std{0.1}} & \textbf{27.3\std{0.4}} & 8.3\std{0.1} & \textbf{98.3\std{0.1}} & \textbf{63.1\std{0.1}} & \textbf{18.6\std{0.1}} & 13.2\std{0.1} & \textbf{77.4\std{0.1}} & \textbf{24.4\std{0.1}} & 15.6\std{0.2} & 43.4\std{0.3} \\
\midrule
\multirow{7}{*}{\rotatebox[origin=c]{90}{bge-m3}}
 & baseline  & 94.9 & 72.9 & 12.6 & 8.8 & 95.6 & 67.1 & 21.2 & 15.3 & 82.1 & 27.7 & 8.1 & 25.4 \\
 & random    & 94.9\std{0.1} & 72.9\std{0.1} & 12.6\std{0.1} & 8.8\std{0.1} & 95.6\std{0.1} & 67.1\std{0.1} & 21.2\std{0.1} & 15.3\std{0.0} & 82.2\std{0.1} & 27.7\std{0.1} & 8.1\std{0.1} & 25.2\std{0.2} \\
 & permuted  & 95.2\std{0.1} & 72.6\std{0.1} & 16.4\std{0.2} & 10.8\std{0.1} & 97.8\std{0.1} & 69.3\std{0.0} & 24.2\std{0.1} & 21.9\std{0.1} & 84.8\std{0.1} & 33.7\std{0.0} & 11.1\std{0.1} & 31.1\std{0.1} \\
 & subspace  & 94.6\std{0.3} & 72.2\std{0.2} & 12.6\std{0.2} & 8.8\std{0.1} & 95.0\std{0.1} & 66.1\std{0.1} & 21.2\std{0.2} & 15.1\std{0.1} & 81.9\std{0.3} & 26.9\std{0.1} & 7.6\std{0.1} & 24.9\std{0.2} \\
 & rotate    & 87.4\std{1.0} & 60.9\std{1.4} & 11.2\std{1.1} & 8.9\std{0.7} & 94.7\std{0.5} & 67.4\std{0.6} & 18.9\std{0.8} & \textbf{15.0\std{0.7}} & 80.0\std{0.8} & 25.9\std{0.8} & 5.4\std{0.5} & 25.6\std{1.9} \\
 & centering & \textbf{95.1\std{0.1}} & \textbf{74.4\std{0.1}} & 13.5\std{0.2} & 9.9\std{0.1} & \textbf{97.8\std{0.1}} & \textbf{71.4\std{0.1}} & 20.2\std{0.1} & 17.8\std{0.1} & \textbf{85.7\std{0.1}} & \textbf{35.3\std{0.0}} & 6.1\std{0.0} & 29.8\std{0.1} \\
 & steering  & 94.7\std{0.2} & 74.1\std{0.1} & \textbf{9.4\std{0.3}} & \textbf{8.6\std{0.1}} & 95.0\std{0.2} & 70.0\std{0.1} & \textbf{12.1\std{0.4}} & 16.3\std{0.3} & 82.6\std{0.2} & 28.2\std{0.1} & \textbf{4.8\std{0.2}} & \textbf{24.8\std{0.3}} \\
\midrule
\multirow{7}{*}{\rotatebox[origin=c]{90}{Qwen3-0.6B}}
 & baseline  & 90.8 & 61.5 & 27.2 & 16.2 & 94.1 & 56.8 & 33.9 & 18.0 & 65.7 & 20.6 & 25.3 & 23.8 \\
 & random    & 90.8\std{0.1} & 61.5\std{0.1} & 27.2\std{0.1} & 16.3\std{0.1} & 94.2\std{0.1} & 56.7\std{0.1} & 33.9\std{0.1} & 18.1\std{0.1} & 65.6\std{0.2} & 20.6\std{0.0} & 25.2\std{0.1} & 23.9\std{0.1} \\
 & permuted  & 86.0\std{0.2} & 50.0\std{0.3} & 54.2\std{0.4} & 9.5\std{0.1} & 90.8\std{0.1} & 50.6\std{0.1} & 48.6\std{0.2} & 17.3\std{0.1} & 61.0\std{0.2} & 18.1\std{0.1} & 47.7\std{0.3} & 16.5\std{0.2} \\
 & subspace  & 91.8\std{0.4} & 66.0\std{0.5} & 19.0\std{1.1} & 12.9\std{0.4} & 94.0\std{0.2} & 57.6\std{0.7} & 31.5\std{1.3} & 16.8\std{0.8} & 66.0\std{0.6} & 20.6\std{0.3} & 22.2\std{1.2} & 23.8\std{0.9} \\
 & rotate    & 82.8\std{1.0} & 50.8\std{1.4} & 21.2\std{2.2} & 14.7\std{1.3} & 92.1\std{0.3} & 55.6\std{0.9} & 30.2\std{1.2} & 18.4\std{1.2} & 59.7\std{0.9} & 16.9\std{0.4} & 10.8\std{1.3} & 23.6\std{2.5} \\
 & centering & \textbf{92.3\std{0.1}} & 67.8\std{0.1} & 18.1\std{0.2} & \textbf{9.2\std{0.1}} & \textbf{96.1\std{0.1}} & 64.9\std{0.0} & 21.9\std{0.1} & \textbf{15.1\std{0.0}} & \textbf{73.2\std{0.1}} & \textbf{26.5\std{0.0}} & 11.1\std{0.1} & \textbf{13.6\std{0.1}} \\
 & steering  & 92.1\std{0.2} & \textbf{68.8\std{0.2}} & \textbf{7.8\std{0.1}} & 13.6\std{0.4} & 95.6\std{0.1} & \textbf{66.8\std{0.2}} & \textbf{11.2\std{0.1}} & 16.3\std{0.2} & 69.7\std{0.1} & 22.8\std{0.1} & \textbf{9.4\std{0.1}} & 23.8\std{0.2} \\
\midrule
\multirow{7}{*}{\rotatebox[origin=c]{90}{Qwen3-8B}}
 & baseline  & 97.6 & 72.7 & 30.8 & 10.4 & 97.0 & 65.9 & 36.9 & 15.6 & 78.3 & 27.9 & 16.1 & 32.3 \\
 & random    & 97.6\std{0.1} & 72.7 & 30.8 & 10.4 & 97.0\std{0.0} & 65.9\std{0.0} & 36.9\std{0.0} & 15.5\std{0.0} & 78.3\std{0.1} & 27.9\std{0.0} & 16.1\std{0.0} & 32.2\std{0.0} \\
 & permuted  & 94.2\std{0.2} & 62.9\std{0.4} & 51.0\std{0.5} & 7.5\std{0.1} & 92.0\std{0.1} & 53.5\std{0.1} & 55.8\std{0.2} & 12.3\std{0.1} & 75.3\std{0.2} & 25.8\std{0.1} & 35.4\std{0.3} & 31.5\std{0.4} \\
 & subspace  & 98.2\std{0.3} & 77.9\std{1.0} & 22.6\std{2.0} & 9.8\std{0.7} & 98.1\std{0.2} & 69.9\std{0.6} & 32.4\std{1.6} & 15.5\std{0.9} & 81.8\std{0.5} & 31.8\std{0.5} & 13.8\std{0.8} & 35.1\std{1.1} \\
 & rotate    & 88.3\std{1.5} & 52.0\std{2.1} & 22.7\std{3.3} & 12.3\std{2.5} & 96.1\std{0.3} & 66.2\std{0.8} & 28.4\std{1.1} & 16.6\std{1.4} & 75.3\std{0.8} & 24.0\std{0.5} & 5.6\std{0.6} & 34.4\std{3.2} \\
 & centering & \textbf{98.9\std{0.1}} & 82.9\std{0.1} & 17.5\std{0.3} & \textbf{8.6\std{0.1}} & 99.2\std{0.0} & 77.3\std{0.1} & 21.0\std{0.1} & \textbf{13.8\std{0.0}} & \textbf{84.5\std{0.1}} & \textbf{35.5\std{0.1}} & 6.8\std{0.1} & \textbf{26.7\std{0.2}} \\
 & steering  & 98.8\std{0.1} & \textbf{83.9\std{0.3}} & \textbf{7.7\std{0.5}} & 10.0\std{0.5} & \textbf{99.3\std{0.0}} & \textbf{79.4\std{0.1}} & \textbf{10.1\std{0.3}} & 14.8\std{0.4} & 81.5\std{0.1} & 30.4\std{0.1} & \textbf{5.2\std{0.2}} & 32.9\std{0.3} \\
\midrule
\multirow{7}{*}{\rotatebox[origin=c]{90}{e5-mistral}}
 & baseline  & 74.3 & 32.2 & 75.3 & 5.3 & 66.5 & 28.6 & 79.5 & 7.5 & 46.3 & 10.4 & 79.2 & 10.7 \\
 & random    & 74.3\std{0.1} & 32.2\std{0.0} & 75.3\std{0.0} & 5.3\std{0.0} & 66.5\std{0.1} & 28.6\std{0.0} & 79.5\std{0.0} & 7.5\std{0.0} & 46.3\std{0.1} & 10.4\std{0.0} & 79.2\std{0.0} & 10.7\std{0.0} \\
 & permuted  & 51.6\std{0.5} & 15.7\std{0.2} & 90.2\std{0.1} & 8.0\std{0.0} & 43.1\std{0.3} & 11.7\std{0.1} & 93.3\std{0.1} & 13.3\std{0.1} & 19.5\std{0.3} & 3.2\std{0.1} & 94.7\std{0.1} & 6.0\std{0.1} \\
 & subspace  & 89.5\std{1.3} & 57.2\std{1.9} & 48.5\std{3.5} & 6.1\std{0.7} & 82.3\std{3.0} & 41.0\std{2.2} & 68.1\std{2.8} & 6.5\std{0.8} & 58.5\std{2.0} & 16.0\std{0.8} & 69.8\std{1.8} & 17.0\std{1.4} \\
 & rotate    & 73.2\std{3.4} & 32.1\std{3.1} & 42.0\std{8.4} & \textbf{2.2\std{0.7}} & 87.9\std{1.4} & 48.7\std{3.1} & 43.1\std{6.5} & \textbf{4.8\std{0.7}} & 65.0\std{1.7} & 17.7\std{1.1} & \textbf{10.2\std{3.6}} & \textbf{6.3\std{1.8}} \\
 & centering & \textbf{94.4\std{0.1}} & \textbf{68.3\std{0.2}} & 32.3\std{0.5} & 6.5\std{0.1} & 97.1\std{0.0} & \textbf{64.9\std{0.1}} & 38.4\std{0.2} & 11.5\std{0.1} & \textbf{82.3\std{0.1}} & \textbf{35.6\std{0.1}} & 20.6\std{0.3} & 19.1\std{0.3} \\
 & steering  & 94.3\std{0.1} & 57.7\std{0.3} & \textbf{20.0\std{0.4}} & 7.0\std{0.1} & \textbf{97.5\std{0.1}} & 59.9\std{0.2} & \textbf{12.0\std{0.1}} & 7.1\std{0.2} & 76.6\std{0.2} & 25.3\std{0.2} & 24.6\std{1.7} & 26.5\std{0.3} \\
\midrule
\multirow{7}{*}{\rotatebox[origin=c]{90}{Gemma-300M}}
 & baseline  & 77.8 & 41.8 & 28.2 & 18.2 & 84.9 & 49.4 & 24.1 & 30.1 & 36.9 & 5.9 & 29.9 & 36.4 \\
 & random    & 77.8\std{0.3} & 41.8\std{0.1} & 28.3\std{0.2} & 18.1\std{0.2} & 85.0\std{0.1} & 49.3\std{0.1} & 24.0\std{0.2} & 30.3\std{0.2} & 36.8\std{0.3} & 5.9\std{0.1} & 29.8\std{0.3} & 36.5\std{0.2} \\
 & permuted  & 51.8\std{0.9} & 18.6\std{0.4} & 80.0\std{0.7} & 9.1\std{0.1} & 79.0\std{0.3} & 37.6\std{0.2} & 56.9\std{0.4} & 24.1\std{0.4} & 14.7\std{0.2} & 1.9\std{0.0} & 75.0\std{0.5} & 15.0\std{0.2} \\
 & subspace  & 79.9\std{0.4} & 45.0\std{0.5} & 19.2\std{1.2} & 19.1\std{0.9} & \textbf{91.8\std{0.3}} & 58.4\std{0.2} & 23.9\std{0.4} & 19.5\std{0.2} & \textbf{43.1\std{3.5}} & \textbf{8.0\std{1.7}} & 21.9\std{4.2} & 40.1\std{3.4} \\
 & rotate    & 46.4\std{1.6} & 17.2\std{0.8} & 17.0\std{2.7} & 22.4\std{2.5} & 71.6\std{0.8} & 37.3\std{0.7} & 18.0\std{0.9} & 29.8\std{1.3} & 21.3\std{0.9} & 2.4\std{0.1} & \textbf{10.0\std{0.8}} & 44.5\std{4.6} \\
 & centering & 79.5\std{0.6} & 41.0\std{0.6} & 39.2\std{1.5} & \textbf{4.7\std{0.2}} & 87.2\std{0.2} & 52.1\std{0.2} & 29.1\std{0.5} & \textbf{8.4\std{0.1}} & 36.1\std{0.3} & 5.6\std{0.0} & 20.5\std{0.6} & \textbf{12.9\std{0.3}} \\
 & steering  & \textbf{81.6\std{0.3}} & \textbf{48.7\std{0.3}} & \textbf{8.3\std{0.7}} & 18.1\std{0.6} & 87.8\std{0.3} & \textbf{59.3\std{0.2}} & \textbf{8.5\std{0.3}} & 21.8\std{0.6} & 42.4\std{0.2} & 7.6\std{0.1} & 9.2\std{0.3} & 39.5\std{0.3} \\
\midrule
\multirow{7}{*}{\rotatebox[origin=c]{90}{\textit{Average}}}
 & baseline  & 86.4 & 52.5 & 44.8 & 10.8 & 87.2 & 50.9 & 48.5 & 16.4 & 64.5 & 18.8 & 37.0 & 27.7 \\
 & random    & 86.4 & 52.4 & 44.8 & 10.8 & 87.2 & 50.9 & 48.4 & 16.4 & 64.4 & 18.8 & 37.1 & 27.6 \\
 & permuted  & 60.8 & 32.6 & 69.5 & 9.3 & 61.3 & 32.4 & 68.1 & 17.5 & 39.5 & 12.2 & 64.5 & 16.8 \\
 & subspace  & 91.3 & 62.5 & 30.3 & 10.1 & 93.0 & 57.9 & 40.4 & \textbf{13.4} & 68.3 & 21.2 & 31.5 & 26.8 \\
 & rotate    & 76.1 & 41.2 & 24.6 & 14.1 & 90.1 & 56.3 & 29.4 & 19.0 & 62.2 & 17.0 & \textbf{10.7} & 36.2 \\
 & centering & 91.2 & 61.6 & 32.1 & \textbf{7.9} & 95.6 & 64.3 & 30.8 & 14.9 & 70.6 & \textbf{26.0} & 15.7 & \textbf{19.3} \\
 & steering  & \textbf{93.1} & \textbf{64.8} & \textbf{14.8} & 10.8 & \textbf{96.1} & \textbf{67.2} & \textbf{12.4} & 15.0 & \textbf{73.6} & 24.4 & \textbf{11.0} & 34.9 \\
\bottomrule
\end{tabular}
\caption{Interventions: cross-lingual Recall@20, nDCG@20, own-language share (Own) and English share (En), averaged over query languages and five estimation seeds ($\pm$ std; omitted when 0.0). Belebele directions are estimated on TED2020, XQuAD and Multi-EuP on FLORES-200. \emph{random} and \emph{permuted} are controls. The best per model and dataset is bold (highest Rec/nDCG, lowest Own/En). The final block averages across models.}
\label{tab:interventions}
\end{table*}

\subsection{What Replaces the Lost Documents}
\label{sec:replacement}

We showed that mixing demotes correct foreign-language documents rather than disordering them. We now ask what takes their place. 
For every query whose cross-lingual gold document leaves the top~$K$ once the index is mixed, we classify each newly entering non-relevant document by its language relative to the query: the query's own language, English, or another language. We also record how often an entrant is a hub; a document is a hub if its retrieval frequency, the number of queries whose top-$K$ list it enters, exceeds the 99th percentile of that distribution. Table~\ref{tab:replacement} reports the composition of the two parallel collections.

\noindent
\textbf{Two regimes, separated by the language of the intruders.} The models fall into two groups. In the first, on Belebele, entrants are almost entirely in the query's own language: $88$--$92\%$ same-language for mE5-base, mE5-large, and e5-mistral, with English essentially absent (at most $0.5\%$). These models bury a query under its own language. In the second group, the same-language material is a minority, and the displaced slots are filled from other languages. For BGE-M3, Qwen, and Gemma, the same-language share falls to $9$--$38\%$ while other-language documents dominate. XQuAD shows the same two regimes. The two regimes seen in the aggregate rankings thus correspond to two distinct sources of interference at the level of individual documents.


\noindent
\textbf{English is the dominant intruder in the second regime.} Within the second regime, English enters far more often than any single other language, rising from near zero in the self-crowding models to $7$--$19\%$ of entrants on Belebele. The other-language column appears larger, but it aggregates every remaining language, eight on Belebele and four on XQuAD, whereas English is a single language. Divided by the number of languages it spans, the other-language share falls to a few percent per language, below the English share for every model except bge-m3. 
The second regime is therefore not a uniform spread across foreign languages but a specific pull toward English. 

\noindent
\textbf{The cause is not hubness, and the drop is late.} Hub documents account for under $9\%$ of entrants in every model, so the demotion is not explained by a few globally attractive vectors dominating nearest-neighbor search. The drop is also a late-layer event: across models, the average layer at which a gold leaves the top list is the last or second-to-last, indicating that the burying is produced by the final task-specializing layers rather than accumulated through the network.

\subsection{Interventions}
\label{sec:interventions-results}

The diagnosis shows that a language's queries are misplaced in the shared space, not that the model fails to represent it. We therefore test corrections that act only on the final representation, comparing the disagreement-guided steering against \emph{mean removal}, \emph{subspace removal}, and \emph{rotation toward a cross-lingual consensus}. All directions and hyperparameters are estimated on FLORES-200 (XQuAD, Multi-EuP) and TED2020 (Belebele).
Table~\ref{tab:interventions} reports recall, nDCG, and the own-language and English shares of the ranking. 

We include two controls: \emph{random}, which removes the same number of directions as subspace removal but does so at random, and \emph{permuted}, which subtracts the mean of a mismatched language. All directions are estimated over five independent estimation corpora samples, and we report the mean and standard deviation across these samples, which measure how much each correction depends on the particular estimation sample rather than on language identity. 

\noindent
\textbf{Every correction recovers recall, but not in the same way.} Mean removal, subspace removal, and steering raise cross-lingual recall over the untouched baseline, and differ in what else moves. Mean removal and steering recover recall while cutting the own-language share that drives the crowding: on mE5-large/Belebele steering lowers it from $69\%$ to $23\%$ and mean removal to $44\%$. Where rotation does help, it does so by pulling rankings toward English, so its English share climbs, most visibly on mE5-large/Multi-EuP, where it rises from $35\%$ to $70\%$, reproducing the bias the correction is meant to remove. Subspace removal is the mildest, moving every quantity only slightly. We report paired significance tests in Appendix~\ref{app:significance}. We note that our design choices are not load-bearing: the rank-fusion constant and the weight $w_\ell(x)$ do not affect the results (Appendix~\ref{app:ablation}).

\noindent
\textbf{Mean removal and steering trade off.} Steering gives the higher recall and the lower own-language share, de-crowding more aggressively: on the same cell, steering drives the own-language share to $8\%$ against $18\%$ for mean removal, at recall $98.8$. Steering also holds up on Multi-EuP, the non-parallel collection, improving recall on the mE5 models, whereas mean removal degrades mE5-base there, where its per-language means are least reliable. Between the two effective corrections, the choice is a trade-off rather than a ranking. On the parallel collections, steering gives the higher nDCG: on Qwen3-8B/Belebele it reaches $83.9$ against $82.9$ for mean removal. Mean removal leads instead to the non-parallel Multi-EuP and in cross-lingual rank agreement, so its rankings are better ordered where steering’s gains are smallest. Appendix~\ref{app:perlang} breaks these effects down by language, showing that both corrections help the lowest-scoring languages most. Steering’s direction is related to but not reducible to the language-mean offset; we analyze this in Appendix~\ref{app:steering-mean}. 


\noindent
\textbf{Controls establish specificity.} Two controls confirm that the corrections act on language identity rather than on the geometry of the space in general. Removing an equal number of random directions, in place of the language subspace, leaves retrieval within noise of the baseline, so the effect of subspace removal is not merely a consequence of discarding dimensions. Subtracting a mismatched language's mean, drawn by permuting the language-to-mean assignment, sharply worsens retrieval and drives the own-language share toward saturation, so it is the correct per-language mean that matters, not the act of centering itself. The corrections that work therefore do so because they target language identity specifically. Appendix~\ref{app:case} traces two individual queries through each correction, illustrating how steering de-crowds the ranking while preserving the same-language answer.

\section{Conclusion}
We studied multilingual dense retrieval through cross-lingual disagreement: how the rankings different languages produce for the same need diverge in isolation and under competition in a shared index. This view separates failures that look identical under relevance metrics. A language may retrieve well alone, preserve its internal ranking under mixing, and still fail in the shared index because its relevant foreign-language documents are demoted beneath the same-language or English material. The failure is thus one of interference, not absent representation: correct cross-lingual documents are pushed down rather than lost, replaced mostly by non-relevant material whose language follows model-dependent regimes.

The same signal also corrects. Among training-free interventions estimated from an external corpus, mean removal, subspace removal, rotation, and disagreement-guided steering repair different aspects of the failure: mean removal gives stronger ranking quality, while steering improves recall and reduces own-language crowding. Relevance, agreement, and language composition move separately, so multilingual retrieval should be judged not only by how many relevant documents are found but by how languages disagree about what to retrieve. Cross-lingual disagreement offers both a label-free diagnostic of this interference and a practical signal for inference-time correction.


\section*{Limitations}

Our analysis treats retrieval disagreement as a diagnostic signal, not as an objective that should always be minimized. Different languages may legitimately retrieve different documents when corpora are not fully parallel, when relevance judgments are incomplete, or when the translated queries emphasize different aspects of the same information need. This is especially important for Multi-EuP, whose documents are not translation-equivalent and whose relevance judgments are graded rather than one-to-one. We therefore interpret disagreement on Multi-EuP as evidence of behavior under realistic multilingual competition, not as a direct semantic mismatch in the same sense as on Belebele and XQuAD.


The interventions require language identifiers and an external multilingual estimation corpus. We use FLORES to avoid fitting directions on the evaluation index, but this assumes that the language offsets and disagreement directions estimated from FLORES transfer to the target domain. Domain mismatch, noisy language identification, code-switching, and languages absent from the estimation corpus may reduce reliability. The steering method is training-free, but it still requires estimating one direction per language and choosing a scale; these choices may not be optimal for all models, domains, or retrieval depths.

We evaluate single-vector dense retrieval at the final embedding layer. We do not study sparse-dense hybrid retrieval, rerankers, late-interaction models, or full retrieval-augmented generation pipelines, where later components may compensate for or amplify the disagreements observed here. We also focus on top-20 retrieval, which is appropriate for common retrieval and RAG settings but does not fully characterize deeper ranking behavior.

Finally, our interventions are diagnostic rather than definitive fixes. Mean removal, subspace removal, rotation, and disagreement-guided steering reveal different aspects of the geometry, but none removes multilingual retrieval bias uniformly. In particular, improving recall, improving nDCG, reducing own-language crowding, and increasing cross-language agreement are not the same objective. This trade-off is part of our finding, but it also limits how directly any single correction can be deployed as a universal solution.


\section*{Ethics Statement}
\label{app:risks}
This work mitigates a fairness-relevant failure, the unequal treatment of query languages in multilingual retrieval, and its gains fall most on the lowest-served languages. Two caveats: our steering correction can raise the English share of results on non-parallel data, which a language-sensitive deployment should weigh; and it assumes the query language is known, so noisy language identification could redistribute rather than remove bias. All datasets are public and contain no sensitive information. The environmental cost is low: the method trains nothing and runs in minutes on cached embeddings.

\section*{Acknowledgements}
The work was supported by the European Research Council (ERC) through the European Union's Horizon Europe research and innovation program (grant agreement No. 101113091) and the German Research Foundation (DFG; grant FR 2829/7-1).

\nocite{ghorbanpour-etal-2025-data, ghorbanpour2026latentspaceinterventioncrosslingualfactual}

\bibliography{custom}
\bibliographystyle{acl_natbib}

\appendix

\section{Datasets: Languages, Access, and Licensing}
\label{app:datasets}

We accessed all datasets from their public Hugging Face repositories and used them solely for research evaluation, consistent with each dataset's intended use.

\textbf{Belebele} \citep{bandarkar-etal-2024-belebele} is a parallel reading comprehension benchmark with 900 questions per language. We use ten languages spanning high- and mid-resource levels and five scripts: English, Spanish, French, Italian, Portuguese, Russian, Arabic, Hindi, Chinese, and Japanese. It is released under CC-BY-SA-4.0, which permits research use with attribution.

\textbf{XQuAD} \citep{Artetxe:etal:2019} is a parallel question answering benchmark of 240 passages and 1{,}190 question--answer pairs per language, translated from SQuAD~v1.1. We use six languages, again spanning high- and mid-resource levels: English,
Spanish, Russian, Arabic, Hindi, and Chinese. It is released under CC-BY-SA-4.0, permitting research use with attribution.

\textbf{Multi-EuP} \citep{yang-etal-2023-multi-eup} is a non-parallel collection from European Parliament debates covering all 24 official EU languages, which range from high-resource (English, German, French) to lower-resource official languages (Maltese, Estonian, Latvian, Lithuanian).
Its imbalanced pools and binary relevance make it our realistic stress test. Derived from public parliamentary proceedings, it is released for non-commercial research, under which our use falls.

\textbf{FLORES-200} \citep{nllbteam2022languageleftbehindscaling} is a parallel corpus covering 200 languages. We use it to estimate the language
statistics for XQuAD and Multi-EuP, taking 500 sentences per language from the development split for the languages present in each set. It is released under ODC-By, which permits research use with attribution.

\textbf{TED2020} \citep{ted_talk} is a corpus of TED and TED-X talk subtitles, crowd-translated into over 100 languages and sentence-aligned across them; this broad, many-way parallel coverage is what lets it serve as an estimation corpus. It is distributed under the TED Talks Usage Policy, under which the talks are licensed CC~BY--NC--ND~4.0, permitting the non-commercial research use.

Belebele comprises 9{,}000 queries over a 4{,}880-document index (10 languages), XQuAD 7{,}140 queries over 1{,}440 documents (6 languages),
and Multi-EuP 22{,}504 queries over 25{,}883 documents (24 languages). For the parallel collections, each query has a single cross-lingual gold per language; on Multi-EuP, the number of relevant documents varies (median: 23, range: 1--188). To our knowledge, none of the datasets contain private or sensitive personal information, and our method requires no additional data collection.

\section{Model and Experimental Details}
\label{app:models}
\noindent
\textbf{Models.} We evaluate seven publicly available embedding models, all accessed through their Hugging Face repositories and run in inference only, without fine-tuning. Table~\ref{tab:models} summarizes their provenance, size, architecture, multilingual status, and license. All are explicitly multilingual and intended for multilingual retrieval, with the exception of e5-mistral, which is trained primarily on English data but is widely used for multilingual retrieval and included here to test whether our findings extend to a decoder not optimized for cross-lingual use.

We read out each model in the pooling and prompt configuration, its documentation specifies: mean pooling with the official \texttt{query:}/\texttt{passage:} prefixes for the mE5 models, CLS pooling for bge-m3, last-token pooling for the Qwen and e5-mistral decoders, and mean pooling with the projection heads applied for embedding-gemma. All licenses permit research use. For the layer-wise analysis, we extract a representation at every layer by applying each model's deployed pooling to that layer's hidden states and run retrieval on the pooled vectors. The pooling is identical across layers, so differences between layers reflect the representations, not the readout.

\noindent
\textbf{Hyperparameters.} All retrieval uses cosine similarity over the final-layer embedding with a cutoff of $K=20$. 
%
All correction hyperparameters are selected on the estimation corpora, and each is set to the value that performs best there: the steering scale $\alpha$, the subspace rank $r$ for subspace removal, and the shrinkage weight for mean removal. The subspace rank peaks at $r=10$ for both a representative encoder and decoder, so no setting advantages our method over the baselines. The steering scale $\alpha$ is tuned per model rather than shared across models, since the architectures tolerate different amounts of steering. Disagreement-guided steering ($\alpha$) is in the range $0.1$--$0.3$, norm-scaled by each query's magnitude, with the consensus for each language formed by reciprocal-rank fusion (rank constant 60) over the other languages. 

\noindent
\textbf{Compute and training.} 
Our method trains nothing. Every correction, including disagreement-guided steering, is estimated in closed form from precomputed embeddings: it introduces no learnable parameters, runs no gradient updates, and adds only a per-language vector or matrix applied to queries at inference. The only forward passes through the models are to extract embeddings for documents, queries, and the FLORES estimation corpus, done once per model and cached. All experiments run on a single GPU (NVIDIA A100 40GB); embedding extraction takes minutes to a few hours per model, depending on size, dominated by the 8B decoders, and every intervention on the cached embeddings, including the full alpha sweep and the multi-seed runs, completes in minutes. The per-language steering directions are the same dimension as the embedding (up to 4096) and add negligible cost at retrieval time.

\begin{table*}[t]
\centering
\small
\setlength{\tabcolsep}{4pt}
\begin{tabular}{l l r l c l}
\toprule
Model & Source & Params & Architecture & Multiling. & License \\
\midrule
mE5-base       & intfloat/multilingual-e5-base   & 278M & encoder & yes & MIT \\
mE5-large      & intfloat/multilingual-e5-large  & 560M & encoder & yes & MIT \\
bge-m3         & BAAI/bge-m3                      & 568M & encoder & yes & MIT \\
Qwen3-0.6B     & Qwen/Qwen3-Embedding-0.6B       & 0.6B & decoder & yes & Apache-2.0 \\
Qwen3-8B       & Qwen/Qwen3-Embedding-8B         & 8B   & decoder & yes & Apache-2.0 \\
e5-mistral     & intfloat/e5-mistral-7b-instruct & 7B   & decoder & (en) & MIT \\
Gemma-300M     & google/embeddinggemma-300m      & 300M & decoder & yes & Gemma \\
\bottomrule
\end{tabular}
\caption{Embedding models evaluated, all obtained from Hugging Face and used in inference
only. ``Multiling.'' indicates whether the model is explicitly trained and released for
multilingual use; e5-mistral is primarily English-trained but applied multilingually. All
listed licenses permit research use.}
\label{tab:models}
\end{table*}

\begin{table*}[t]
\centering
\small
\setlength{\tabcolsep}{5pt}
\begin{tabular}{c l c p{9.2cm}}
\toprule
\multicolumn{4}{l}{\textbf{Query} (it): \emph{Quale dei seguenti aspetti continuer\`a nonostante la presenza di stimoli esterni?}} \\
\midrule
Rank & Lang & Gold & Document snippet \\
\midrule
\multicolumn{4}{l}{\textit{Baseline}} \\
1 & it & \checkmark & In molti hanno osservato che i ritmi fisiologici e il comportamento dipendono in maniera\ldots \\
2 & it & & La stazione non ha cambiato approccio, nonostante la precedente perdita di un giroscopio\ldots \\
3 & it & & L'avidit\`a e l'egoismo faranno sempre parte di noi. \`E insito nella natura della cooperazione\ldots \\
4 & it & & Internet riunisce in s\'e alcuni elementi sia della comunicazione di massa\ldots \\
5 & it & & A casa, data questa costante esposizione ai germi locali\ldots \\
\midrule
\multicolumn{4}{l}{\textit{Steering}} \\
1 & it & \checkmark & In molti hanno osservato che i ritmi fisiologici e il comportamento dipendono in maniera\ldots \\
2 & ru & \checkmark & russian text \\
3 & it & & La stazione non ha cambiato approccio, nonostante la precedente perdita di un giroscopio\ldots \\
4 & it & & Internet riunisce in s\'e alcuni elementi sia della comunicazione di massa\ldots \\
5 & hi & \checkmark & hindi text \\
\bottomrule
\end{tabular}
\caption{Top-5 retrievals for an Italian query under mE5-large. The baseline returns only
Italian passages and no cross-lingual gold; steering keeps the Italian answer and surfaces
the Russian and Hindi copies.}
\label{tab:case1}
\end{table*}

\begin{table*}[t]
\centering
\small
\setlength{\tabcolsep}{5pt}
\begin{tabular}{c l c p{9.2cm}}
\toprule
\multicolumn{4}{l}{\textbf{Query} (pt): \emph{\ldots qual dos seguintes \'e considerado uma solu\c{c}\~ao de longo prazo vi\'avel para quest\~oes ambientais?}} \\
\midrule
Rank & Lang & Gold & Document snippet \\
\midrule
\multicolumn{4}{l}{\textit{Baseline}} \\
1 & pt & \checkmark & Podemos come\c{c}ar a viver de maneira mais amig\'avel com o ambiente, podemos nos juntar ao movimento\ldots \\
2 & pt & & Outros assuntos na agenda em Bali incluem salvar as florestas remanescentes do mundo\ldots \\
3 & pt & & Na savana, \'e dif\'icil para um primata com um sistema digestivo como o dos humanos\ldots \\
4 & pt & & Peter Costello, um tesoureiro da Austr\'alia e o homem com mais chance de suceder\ldots \\
5 & pt & & A emenda proposta j\'a teve sua aprova\c{c}\~ao nas duas casas em 2011\ldots \\
\midrule
\multicolumn{4}{l}{\textit{Steering}} \\
1 & pt & \checkmark & Podemos come\c{c}ar a viver de maneira mais amig\'avel com o ambiente, podemos nos juntar ao movimento\ldots \\
2 & pt & & Outros assuntos na agenda em Bali incluem salvar as florestas remanescentes do mundo\ldots \\
3 & pt & & Na savana, \'e dif\'icil para um primata com um sistema digestivo como o dos humanos\ldots \\
4 & pt & & Peter Costello, um tesoureiro da Austr\'alia e o homem com mais chance de suceder\ldots \\
5 & es & \checkmark & Podemos comenzar a llevar una vida m\'as amigable con el planeta, formar parte de movimientos\ldots \\
\bottomrule
\end{tabular}
\caption{Top-5 retrievals for a Portuguese query under mE5-large. The baseline returns only
Portuguese passages; steering retains the Portuguese answer and surfaces the Spanish
parallel copy at rank~5.}
\label{tab:case2}
\end{table*}

\section{AI Usage Statement}

The authors used an AI assistant to support coding and writing. For code, it helped draft and debug analysis and evaluation scripts, which the authors reviewed, tested, and verified
against the reported results. For writing, it assisted with phrasing and copy-editing of text drafted by the authors. All research ideas, experimental design, analysis, and conclusions are the authors' own, and the authors take full responsibility for the content of the paper.

\section{Case Studies}
\label{app:case}

To make the mechanism concrete, we trace two Belebele queries under mE5-large through the
baseline, mean removal, and disagreement-guided steering. In both, the gold answer exists as
a parallel passage in every language, so a language-agnostic retriever should surface those
copies rather than filling the ranking with same-language distractors.
Tables~\ref{tab:case1} and~\ref{tab:case2} show the top-5 retrievals with a snippet of each
document. In both examples the baseline returns only same-language passages and recovers no
cross-lingual gold, while steering keeps the same-language answer and surfaces parallel
copies in other languages. Mean removal also recovers cross-lingual golds in the first
example, but at the cost of dropping the same-language passage from the top rank.

\section{Demotion vs.\ Loss: Larger Cutoffs}
\label{app:k100}

If the correct foreign documents were absent rather than demoted, widening the retrieval window would not recover them. It does. Raising the cutoff from $K{=}20$ to $K{=}100$ increases cross-lingual recall by $11$--$21$ points on the parallel collections, while nDCG rises far less, since the recovered documents sit deep in the ranking (Table~\ref{tab:k100}). The correct documents are therefore present and retrievable but
demoted. On the non-parallel Multi-EuP collection, many golds fall beyond rank $100$, consistent with it being the hardest setting.

\begin{table}[t]
\centering
\small
\begin{tabular}{l cc cc}
\toprule
& \multicolumn{2}{c}{Belebele} & \multicolumn{2}{c}{XQuAD} \\
\cmidrule(lr){2-3}\cmidrule(lr){4-5}
Model & R@20 & R@100 & R@20 & R@100 \\
\midrule
mE5-base   & 56 & 77 & 71 & 90 \\
mE5-large  & 63 & 82 & 78 & 93 \\
Qwen3-0.6B & 75 & 88 & 82 & 94 \\
Qwen3-8B   & 84 & 94 & 88 & 97 \\
\bottomrule
\end{tabular}
\caption{Cross-lingual Recall at $K{=}20$ vs.\ $K{=}100$ on the parallel collections.
Demoted golds reappear at the larger cutoff, confirming demotion rather than loss.}
\label{tab:k100}
\end{table}

\section{Design-Choice Ablations}
\label{app:ablation}

Two components of the pipeline could in principle drive the results: the
rank-fusion constant used to combine retrieval lists, and the agreement
weight $w_\ell(x)$ used when estimating the steering direction. Neither does.
Table~\ref{tab:ablation} reports steering nDCG on Belebele for four models
spanning both encoders and decoders, under three RRF constants and with the
weight removed.

\begin{table}[t]
\centering
\small
\begin{tabular}{l cccc}
\toprule
Model & RRF=20 & RRF=60 & RRF=100 & no $w_\ell$ \\
\midrule
mE5-base    & 58.80 & 58.80 & 58.80 & 58.80 \\
mE5-large   & 66.79 & 66.79 & 66.79 & 66.79 \\
Qwen3-0.6B  & 69.22 & 69.22 & 69.21 & 69.22 \\
Qwen3-8B    & 83.66 & 83.67 & 83.66 & 83.67 \\
\bottomrule
\end{tabular}
\caption{Steering nDCG@20 on Belebele under three rank-fusion constants and with the agreement weight removed. Values move by at most $0.01$ across all conditions; cross-lingual recall is identical to within $0.01$. The main results use RRF=60.}
\label{tab:ablation}
\end{table}

The fusion constant and the agreement weight are therefore inert at the scale of our effects: the corrections reported in the main paper reflect the interventions themselves, not tuning of the fusion or weighting machinery.

\section{Significance Testing}
\label{app:significance}

We test steering against mean removal with paired Wilcoxon signed-rank tests over all model$\times$dataset$\times$language cells ($n=184$), per metric. The results confirm that steering's advantage is de-crowding rather than a uniform win on retrieval quality (Table~\ref{tab:significance}). The reduction in own-language share is large and highly significant ($\Delta=-9.7$, $p<10^{-19}$) and holds on all three collections individually. The recall difference favors steering on average ($+2.7$) but is not significant ($p=0.28$), reflecting high per-language variance. nDCG is significant in opposite directions by collection: mean removal is higher on Belebele ($p=0.004$) and steering on XQuAD ($p=0.007$). These tests substantiate the trade-off framing: steering reliably de-crowds the ranking, while recall and nDCG depend on the model and collection.

\begin{table}[t]
\centering
\small
\begin{tabular}{l rr r}
\toprule
Metric & Steering & Mean rem. & $p$ \\
\midrule
Own-lang share ($\downarrow$) & 11.3 & 21.0 & $<10^{-19}$ \\
Recall & 87.4 & 84.7 & $0.28$ \\
nDCG (Belebele) & 65.8 & 67.1 & $0.004$ \\
nDCG (XQuAD) & 67.2 & 64.3 & $0.007$ \\
\bottomrule
\end{tabular}
\caption{Paired Wilcoxon signed-rank tests, steering vs.\ mean removal, over
model$\times$dataset$\times$language cells ($n=184$). Steering's reduction in own-language crowding is highly significant; recall is not significant; nDCG is significant in opposite directions by collection.}
\label{tab:significance}
\end{table}

\section{Per-Language Correction Effects}
\label{app:perlang}

\begin{figure*}[t]
\centering
\includegraphics[width=\textwidth]{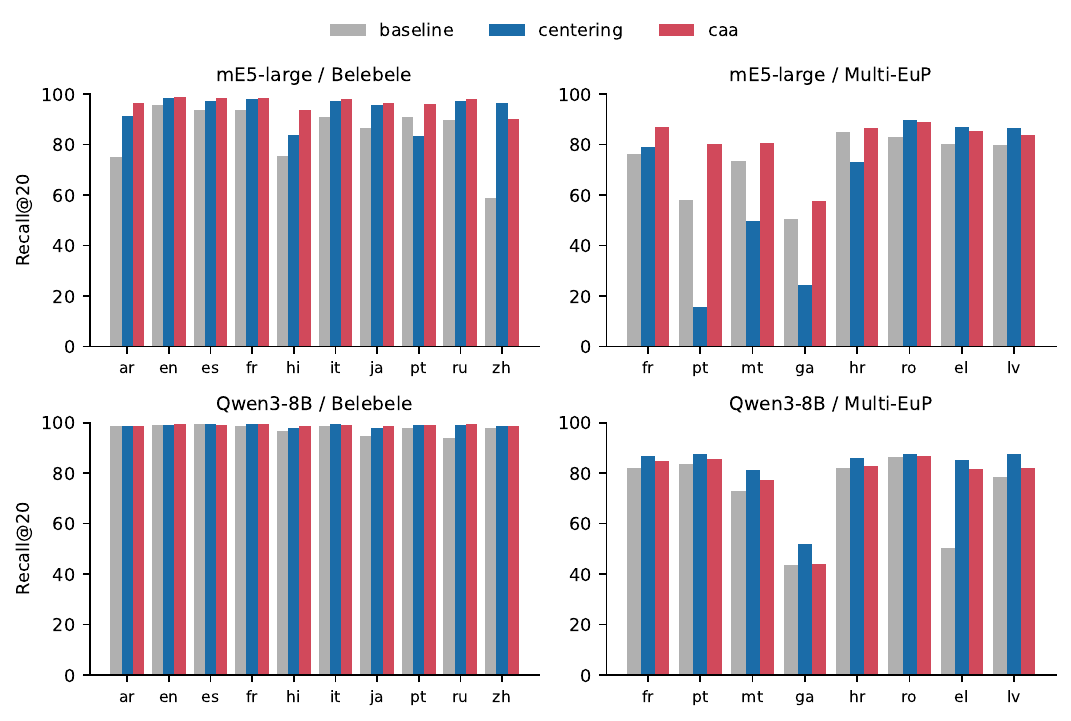}
\caption{Per-language Recall@20 underbaseline, mean removal (centering), and disagreement-guided steering (caa), for mE5-large and Qwen3-8B on Belebele (all ten languages) and a subset of Multi-EuP. The Multi-EuP languages shown are a resource-balanced subset chosen to include cases where each correction wins, not to favor ours.
}
\label{fig:perlang}
\end{figure*}

The main results report intervention effects averaged over query languages, which can hide which languages a correction actually helps. Figure~\ref{fig:perlang} breaks the effect down by language, comparing baseline, mean removal, and disagreement-guided steering on Recall@20 for an encoder (mE5-large) and a decoder (Qwen3-8B), on the parallel Belebele collection and a subset of the non-parallel Multi-EuP.

On Belebele, both corrections lift the lowest-scoring languages most, compressing the spread across languages, and the two behave similarly. The non-parallel Multi-EuP subset separates them sharply and reveals where steering matters. For the encoder, mean removal collapses on several languages, dropping recall to $16$ on Portuguese and $24$ on Irish, precisely the cases where its per-language mean is least reliable on non-parallel data. Steering does not: it restores recall to $80$ on Portuguese and $58$ on Irish, and improves the low-resource Maltese from $50$ to $81$. On the remaining languages, the two are close, with mean removal holding a small edge on a few. Steering thus rescues exactly the languages the alternative abandons.

The decoder behaves differently. Its baseline recall is already high and uniform across languages, so there is little for either correction to recover, and mean removal is marginally better throughout. The per-language view, therefore, mirrors the aggregate trade-off at its extreme: steering is most valuable where the model and setting are hardest, the encoder on non-parallel data, and least needed where retrieval is already strong.

\section{Relationship Between Steering and Mean Removal}
\label{app:steering-mean}

Because the steering direction $v_\ell$ contrasts other-language documents against language $\ell$'s own, it is natural to ask whether it reduces to the language-mean offset $\bar{\mu}-\mu_\ell$ that mean removal subtracts. We find the two are correlated but distinct. The cosine similarity $\cos(v_\ell,\,\bar{\mu}-\mu_\ell)$ is $0.67$ on average across models (ranging from $0.62$ to $0.73$), so the directions overlap substantially, as
expected since language identity dominates the embedding geometry, but roughly half of $v_\ell$'s magnitude is orthogonal to the mean direction.

To test whether this orthogonal component carries a signal, we evaluate \emph{Residual}, a variant that projects $v_\ell$ orthogonal to $\bar{\mu}-\mu_\ell$ before steering. Removing the mean-aligned component degrades both recall and nDCG for every model except bge-m3 (Table~\ref{tab:residual}), with the largest drops on the models where steering helps most (e.g.\ mE5-large nDCG falls from $66.8$ to $50.1$). Steering is therefore not a rescaled mean offset: the behaviorally estimated direction captures structure that a static per-language mean
does not.

\begin{table}[t]
\centering
\small
\begin{tabular}{l cc cc c}
\toprule
& \multicolumn{2}{c}{Steering} & \multicolumn{2}{c}{Residual} & \\
\cmidrule(lr){2-3}\cmidrule(lr){4-5}
Model & Rec & nDCG & Rec & nDCG & $\cos$ \\
\midrule
mE5-large  & 97.2 & 66.8 & 92.1 & 50.1 & 0.64 \\
mE5-base   & 94.3 & 58.8 & 89.6 & 45.3 & 0.62 \\
bge-m3     & 94.5 & 74.8 & 94.9 & 72.7 & 0.73 \\
Qwen3-0.6B & 92.3 & 69.2 & 91.2 & 63.4 & 0.65 \\
Qwen3-8B   & 99.0 & 83.7 & 98.4 & 76.4 & 0.66 \\
Gemma-300M & 82.0 & 49.9 & 77.9 & 42.1 & 0.71 \\
\bottomrule
\end{tabular}
\caption{Steering vs.\ its mean-removed residual, cross-lingual Recall@20 / nDCG@20 on Belebele, and the cosine between the steering direction and the language-mean offset. Removing the mean-aligned component degrades steering for every model but bge-m3, confirming the orthogonal component carries signal.}
\label{tab:residual}
\end{table}

\section{Agreement Under Interventions}
\label{app:agreement}

We report cross-lingual agreement under each correction: set agreement A1 (Jaccard overlap of top-$K$ lists) and rank agreement A2 (Spearman over
shared documents), averaged over languages on Belebele (Table~\ref{tab:agreement}). Both corrections generally raise agreement over the baseline, most on the self-crowding models, where interference was largest; on bge-m3, whose baseline agreement is already high, steering leaves it roughly unchanged. The de-crowding, therefore, reflects convergence toward the cross-lingual consensus rather than added noise.

\begin{table}[t]
\centering\small
\setlength{\tabcolsep}{4pt}
\begin{tabular}{l ccc ccc}
\toprule
& \multicolumn{3}{c}{A1 (set)} & \multicolumn{3}{c}{A2 (rank)} \\
\cmidrule(lr){2-4}\cmidrule(lr){5-7}
Model & Base & Cent & Steer & Base & Cent & Steer \\
\midrule
mE5-base    & 30.4 & 28.2 & 33.1 & 57.9 & 59.8 & 60.5 \\
mE5-large   & 36.4 & 37.9 & 40.1 & 59.7 & 65.2 & 63.4 \\
bge-m3      & 51.9 & 51.9 & 48.6 & 73.6 & 75.0 & 71.1 \\
Qwen3-0.6B  & 39.1 & 43.7 & 39.6 & 64.9 & 70.6 & 62.2 \\
Qwen3-8B    & 33.6 & 44.5 & 41.1 & 67.9 & 74.9 & 68.6 \\
e5-mistral  & 33.5 & 43.4 & 43.0 & 56.4 & 68.3 & 61.3 \\
Gemma-300M  & 25.9 & 23.4 & 26.4 & 42.5 & 47.2 & 41.9 \\
\bottomrule
\end{tabular}
\caption{Cross-lingual agreement under interventions on Belebele, averaged
over languages ($\times100$): set agreement A1 and rank agreement A2 for
baseline, mean removal (Cent), and steering (Steer).}
\label{tab:agreement}
\end{table}
\end{document}